\UseRawInputEncoding
\documentclass[a4paper,10pt,oneside]{article}
\usepackage[utf8]{inputenc}
\usepackage{twomod,amsmath,epsfig,times,url,hyperref}
\usepackage[T1]{fontenc}
\usepackage{booktabs}       
\usepackage{caption}        

\title{Pre-training of Lightweight Vision Transformers on Small Datasets with Minimally Scaled Images}

\author{
 Jen Hong Tan \\
  Data Science and Artificial Intelligence Lab\\
  Health Services Research Unit\\
  Singapore General Hospital \\
}

\begin{document}
\ninept
\maketitle
\begin{sloppy}
\begin{abstract}
Can a lightweight Vision Transformer (ViT) match or exceed the performance of Convolutional Neural Networks (CNNs) like ResNet on small datasets with small image resolutions? This report demonstrates that a pure ViT can indeed achieve superior performance through pre-training, using a masked auto-encoder technique with minimal image scaling. Our experiments on the CIFAR-10 and CIFAR-100 datasets involved ViT models with fewer than 3.65 million parameters and a multiply-accumulate (MAC) count below 0.27G, qualifying them as 'lightweight' models. Unlike previous approaches, our method attains state-of-the-art performance among similar lightweight transformer-based architectures without significantly scaling up images from CIFAR-10 and CIFAR-100. This achievement underscores the efficiency of our model, not only in handling small datasets but also in effectively processing images close to their original scale.
\end{abstract}

\section{Introduction}
\label{sec:intro}

In the rapidly evolving field of computer vision, the quest for more efficient and effective models remains a central research focus. Traditionally, Convolutional Neural Networks (CNNs) \cite{lecun1989backpropagation, krizhevsky2012imagenet, he2016deep} have dominated this domain, especially in tasks involving small datasets. However, the emergence of Vision Transformers (ViTs) \cite{dosovitskiy2020image} has introduced a paradigm shift, challenging the supremacy of CNNs in certain applications. Despite their success in large-scale datasets, ViTs have struggled to match the performance of CNNs in smaller data scenarios due to their lack of inherent inductive biases \cite{dosovitskiy2020image}, as a result, CNNs are still the preferred architectures for smaller datasets \cite{hassani2021escaping}.

Addressing this gap, our study explores the potential of ViTs in contexts where data availability is limited. The central hypothesis of our research is that a lightweight ViT can perform as well as, or even better than, its CNN counterparts like ResNet \cite{he2016deep} on small datasets. To investigate this, we have employed a pre-training strategy using  Masked Auto-Encoder (MAE) \cite{he2022masked} to enhance the capability of ViTs to learn from smaller datasets effectively.

In this report, we detail a series of experiments conducted to test the effectiveness of lightweight ViTs when pre-trained with a MAE on small datasets. We present the adaptations made to the conventional ViT and MAE structure to improve its suitability for smaller datasets and document the implementation of our pre-training regimen. Our results indicate that with these modifications, ViTs can indeed reach or exceed the performance benchmarks set by CNNs on smaller datasets.

\section{Method}
\label{sec:format}

In our implementation of the Masked Auto-Encoder (MAE), we largely follow the process outlined in the original MAE paper \cite{he2022masked}. We start by dividing each image into non-overlapping patches, similar to the ViT \cite{dosovitskiy2020image}. From these patches, a subset is randomly selected without replacement, following a uniform distribution. These selected patches are known as the unmasked patches and fed into the Encoder. The remaining patches are the masked patches, which are not directly used in the encoding step. We adopt the masking ratio of 0.75, as recommended by the original MAE paper, so that 75\% of the total patches are masked and not included in the initial input to the encoder. See Figure \ref{fig:mae_algo}.

\begin{figure*} 
    \includegraphics[trim=0.3cm 5cm 5.7cm 0cm, clip, scale=0.7]{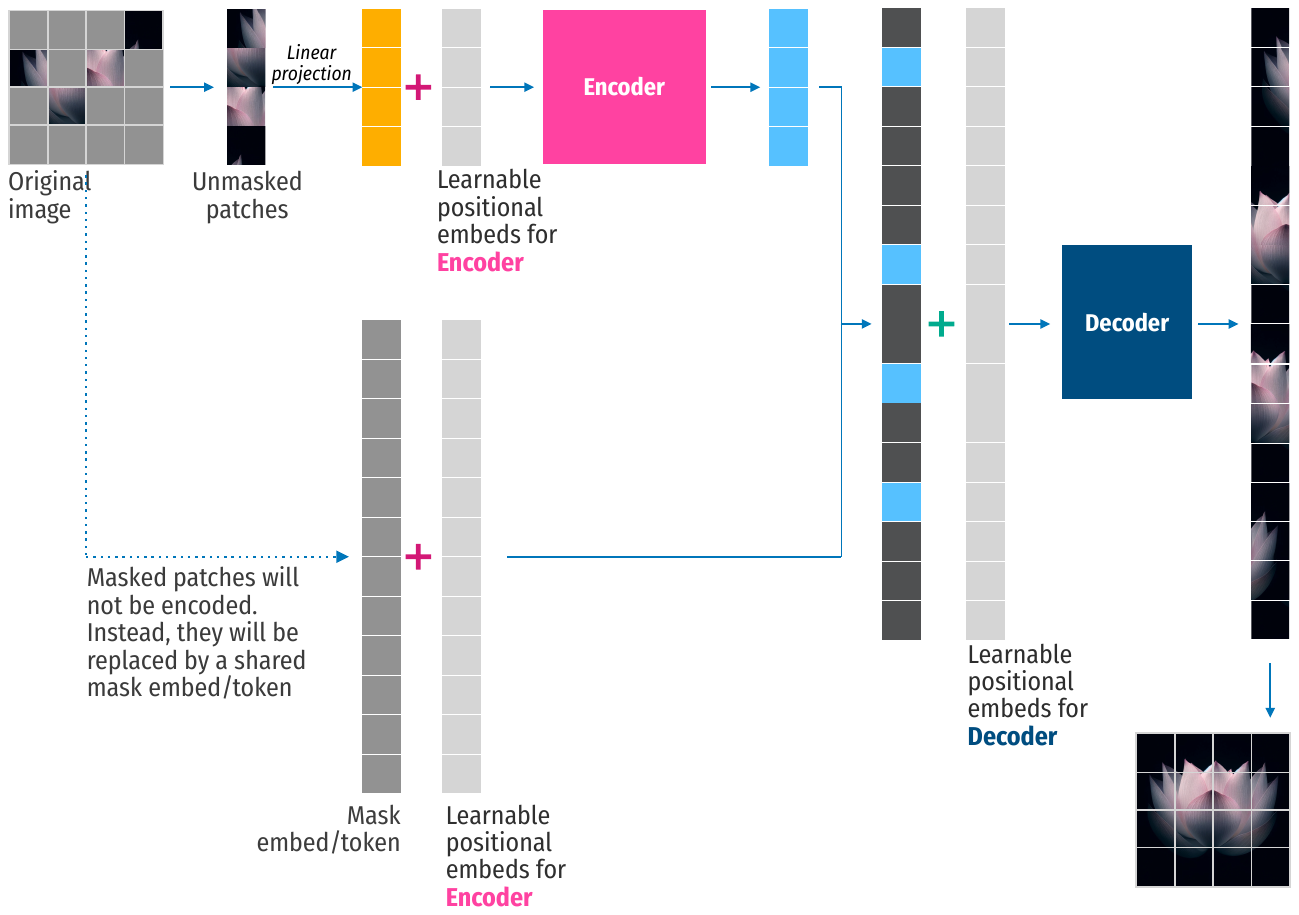}
    \caption{Modified MAE: Implementation using separate learnable positional embeddings for Encoder and Decoder. The Encoder and the Decoder in this figure only consist of transformer layers, a  layer of layerNorm and a linear projection layer.}
    \label{fig:mae_algo}
\end{figure*}

\subsection{MAE Encoder}
\label{ssec:mae_en}
Our encoder adopts the structure of a ViT \cite{dosovitskiy2020image}, but it operates only on the visible, unmasked patches of an image. In line with standard ViT models, our encoder first applies a linear projection to embed these patches, enhancing them with positional information, followed by a series of transformer layers \cite{NIPS2017_attn} and a linear projection layer. Layer normalization \cite{ba2016layer} is applied before the input to the linear projection layer. See Figure \ref{fig:mae_arch}.

However, diverging from both the original ViT and MAE implementations \cite{he2022masked}, our encoder utilizes learnable positional embeddings instead of the traditional sinusoidal positional embeddings. Given that our masking ratio is set to 0.75, the encoder is essentially processing only a quarter (25\%) of the total patches in an image. The masked patches, which comprise the majority of the image data, are excluded from this phase of processing. 

\begin{figure*} 
\centering
    \includegraphics[trim=0.9cm 2cm 3.5cm 0cm, clip, scale=0.6]{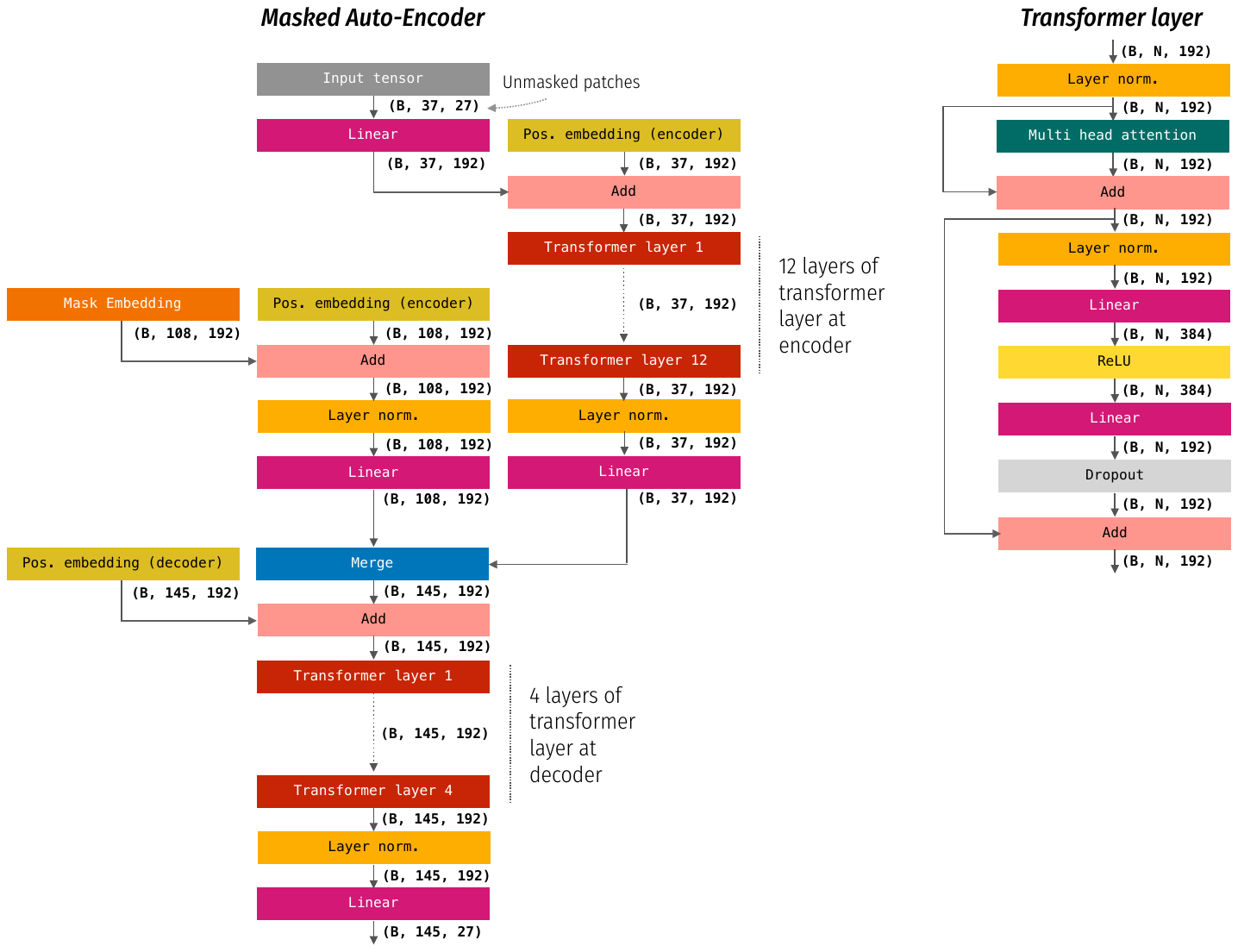}
    \caption{The architecture of our MAE. 'B' stands for batch size, 'N' for number of embeddings}
    \label{fig:mae_arch}
\end{figure*}

\subsection{MAE Decoder}
\label{ssec:mae_de}

The MAE decoder in our model is responsible for reconstructing the image from a comprehensive set of embeddings, which includes both the encoded visible patches and mask embeddings. These mask embeddings \cite{Devlin2019BERTPO} are shared, learnable vectors that signify the presence of each missing patch that the model aims to predict. Unlike the original MAE implementation \cite{he2022masked}, we incorporate separate learnable positional embeddings at the decoder's input, ensuring precise spatial awareness in the reconstruction phase.

Structurally, the decoder consists of a series of transformer layers, a layer normalization and a linear projection layer, similar to the encoder. However, the decoder's architecture is not constrained by the design of the encoder. This independence allows for greater flexibility in its configuration. We can, for instance, opt for a different embedding size or a varying number of transformer layers in the decoder compared to the encoder.

The use of the MAE decoder is exclusive to the pre-training phase, where its primary function is to reconstruct the original image from the partial input provided by the encoder. This specialized role allows us to tailor the decoder's design specifically for optimal reconstruction performance, independent of the encoder's configuration.

\subsection{Reconstruction Output}
\label{ssec:re_out}

In our MAE model, the reconstruction of the input image is achieved by predicting the pixel values for each masked patch, aligning with the approach used in the original MAE. The decoder's output for each patch is a vector that corresponds to the pixel values of the patch. This output undergoes a linear projection in the final layer of the decoder, with the number of output neurons matching the number of pixel values in a patch. We then reshape this output to form the reconstructed image.

A key aspect of our implementation is the application of the loss function. The original MAE uses the mean squared error (MSE) to measure the difference between the reconstructed and original images in pixel space. However, unlike the original MAE where the loss is calculated solely on the masked patches (akin to the strategy used in BERT \cite{Devlin2019BERTPO}), our approach modifies this calculation. We compute the total loss as a sum of the loss on masked patches and an additional, discounted loss on unmasked patches:

\begin{equation}
\mathrm{Loss} = \mathrm{MSE}_{\mathrm{masked\_patches}} + \alpha \cdot \mathrm{MSE}_{\mathrm{unmasked\_patches}}
\label{eq:loss}
\end{equation}

$\alpha$ denotes the discounting factor. This variation addresses a specific issue observed in the original implementation, where no loss was computed on visible patches, leading to a lower quality in model output for these patches. By incorporating a discounted loss on unmasked patches, we not only observe a reduction in the total loss (compared to calculating it only on masked patches) but also a significant improvement in the quality of the reconstructed visible patches.

\section{Experimental Setup}
\label{sec:experiment}

In our experimental setup, we conduct self-supervised pre-training on the CIFAR10 \cite{krizhevsky2009learning} and CIFAR100 \cite{krizhevsky2009learning} training datasets separately. Following this pre-training phase, each model is fine-tuned for classification tasks on the respective CIFAR10 and CIFAR100 datasets. We use the same random seed for all of our experiments.

A notable aspect of our implementation is the adjustment of the image input size to 36 x 36, slightly larger than the original 32 x 32 dimension of these datasets. This size change results in each image being divided into 144 patches, given our chosen patch size of 3 x 3.

We add an auxiliary dummy patch to each image during both the pre-training and fine-tuning phases. This dummy patch, which contains zero values for all elements, is appended as the 145th patch. The encoded embedding of this auxiliary patch is utilized for classification purposes. Consequently, this adjustment necessitates a total of 145 positional embeddings for both the encoder and decoder in our model, accommodating the extra dummy patch.

\subsection{Encoder and Decoder design}

Our Masked Auto-Encoder employs encoders and decoders both with an embedding size of 192, and each utilizes 3 heads. The encoder is composed of 12 transformer layers, while the decoder features 4 transformer layers. 

Within each transformer layer (see Figure \ref{fig:mae_arch}), we opt for the ReLU \cite{Nair2010RectifiedLU} activation function instead of the GELU \cite{hendrycks2023gaussian}, which is used in the original Vision Transformer (ViT) \cite{dosovitskiy2020image}. Each layer includes a sequence of two linear layers: the first expanding the output to twice (\textit{rather than 4 times}) the embedding size, followed by ReLU activation, and the second linear layer reducing it back to the embedding size. We apply dropout \cite{JMLR:v15:srivastava14a} with a rate of 0.1 to the output of these layers . All linear layers within the transformer layers are included with bias. However, we exclude bias from other linear projection layers in both the encoder and decoder.

For initializing weights and biases across all layer types, we rely on the default methods provided by Pytorch. The same applies to our approach to layer normalization \cite{ba2016layer}, where we use Pytorch's default setup.

\subsection{Pre-training}
Our pre-training process is arranged for 4000 epochs, employing the AdamW optimizer \cite{loshchilov2018decoupled} with a weight decay set at 0.05. The initial 400 epochs are designated for warm-up \cite{goyal2018accurate}. We follow this with a cosine decay schedule \cite{loshchilov2017sgdr} for the learning rate. The batch size during training is 1408, and we adhere to the linear learning rate scaling rule with a base learning rate of $1.5e-4$ \cite{goyal2018accurate, he2022masked}:

\begin{equation}
\mathrm{lr} = \mathrm{base\_lr} \times \mathrm{batch\_size} /256
\label{eq:lr}
\end{equation}

In terms of data augmentation, we randomly flip the input images horizontally with a 50\% probability. Additionally, we utilize random resized cropping, with a scale range from 0.6 to 1. For color normalization, we apply a mean of 0.5 and a standard deviation of 0.5 to each color channel. The discounting factor ($\alpha$) in Eq. \ref{eq:lr} is set to 0.1. See Table \ref{tab:pretrain_recipe} for recipe.

\begin{table}
\captionsetup{skip=5pt}
\centering
\caption{Parameters and Configuration for Pre-Training}
\begin{tabular}{ll}
\toprule
\textbf{Configuration} & \textbf{Value} \\
\midrule
Optimizer & AdamW \\
Weight decay & 0.05 \\
Base learning rate & $1.5e-4$ \\
Learning rate schedule & Cosine decay \\
Total epochs & 4000 \\
Warm-up epochs & 400 \\
Batch size & 1408 \\
Horizontal flipping & $p=0.5$ \\
Random resized cropping & $[0.6, 1]$ \\
Color normalization & $\mathrm{mean}=0.5, \mathrm{std}=0.5$\\
Discounting factor ($\alpha$) & 0.1 \\

\bottomrule
\end{tabular}
\label{tab:pretrain_recipe}
\end{table}

\subsection{Fine-tuning}

For fine-tuning, we use the pre-trained MAE encoder for classification tasks on CIFAR10 and CIFAR100 datasets. The fine-tuning architecture comprises the MAE encoder and an additional linear layer. The input to this linear layer is the last embedding from the encoder, which corresponds to the dummy patch. We fine-tune for 300 epochs, with the initial 20 epochs serving as a warm-up period \cite{goyal2018accurate}. This is followed by a cosine decay learning rate schedule \cite{loshchilov2017sgdr}. The optimizer used is AdamW \cite{loshchilov2018decoupled} with a weight decay of 0.05, and the batch size is set at 768. We adhere to a linear learning rate scaling rule (Eq. \ref{eq:loss}), starting with a base learning rate of $1e-3$ and applying a layer-wise decay \cite{ClarkLLM20} at a rate of 0.75.

Data augmentation during fine-tuning includes a random horizontal flip with a 50\% probability and the application of the AutoAugment policy tailored for CIFAR10 \cite{autoaugment} (on both CIFAR10 and CIFAR100). We also perform random resized cropping with a scale range of 0.8 to 1.0. Color normalization is conducted on each color channel, with a mean of 0.5 and a standard deviation of 0.5. See Table \ref{tab:finetune_recipe} for recipe.

\begin{table}
\captionsetup{skip=5pt}
\centering
\caption{Parameters and Configuration for Fine-Tuning}
\begin{tabular}{ll}
\toprule
\textbf{Configuration} & \textbf{Value} \\
\midrule
Optimizer & AdamW \\
Weight decay & 0.05 \\
Base learning rate & $1e-3$ \\
Learning rate schedule & Cosine decay \\
Layer-wise decay & 0.75 \\
Total epochs & 300 \\
Warm-up epochs & 20 \\
Batch size & 768 \\
Horizontal flipping & $p=0.5$ \\
Random resized cropping & $[0.8, 1]$ \\
AutoAugment &Policy for CIFAR-10 \\
Color normalization & $\mathrm{mean}=0.5, \mathrm{std}=0.5$\\

\bottomrule
\end{tabular}
\label{tab:finetune_recipe}
\end{table}

\section{Result}
\label{sec:result}

\begin{figure} 
    \centering
    \includegraphics[trim=0cm 0cm 0cm 0cm, clip, scale=0.19]{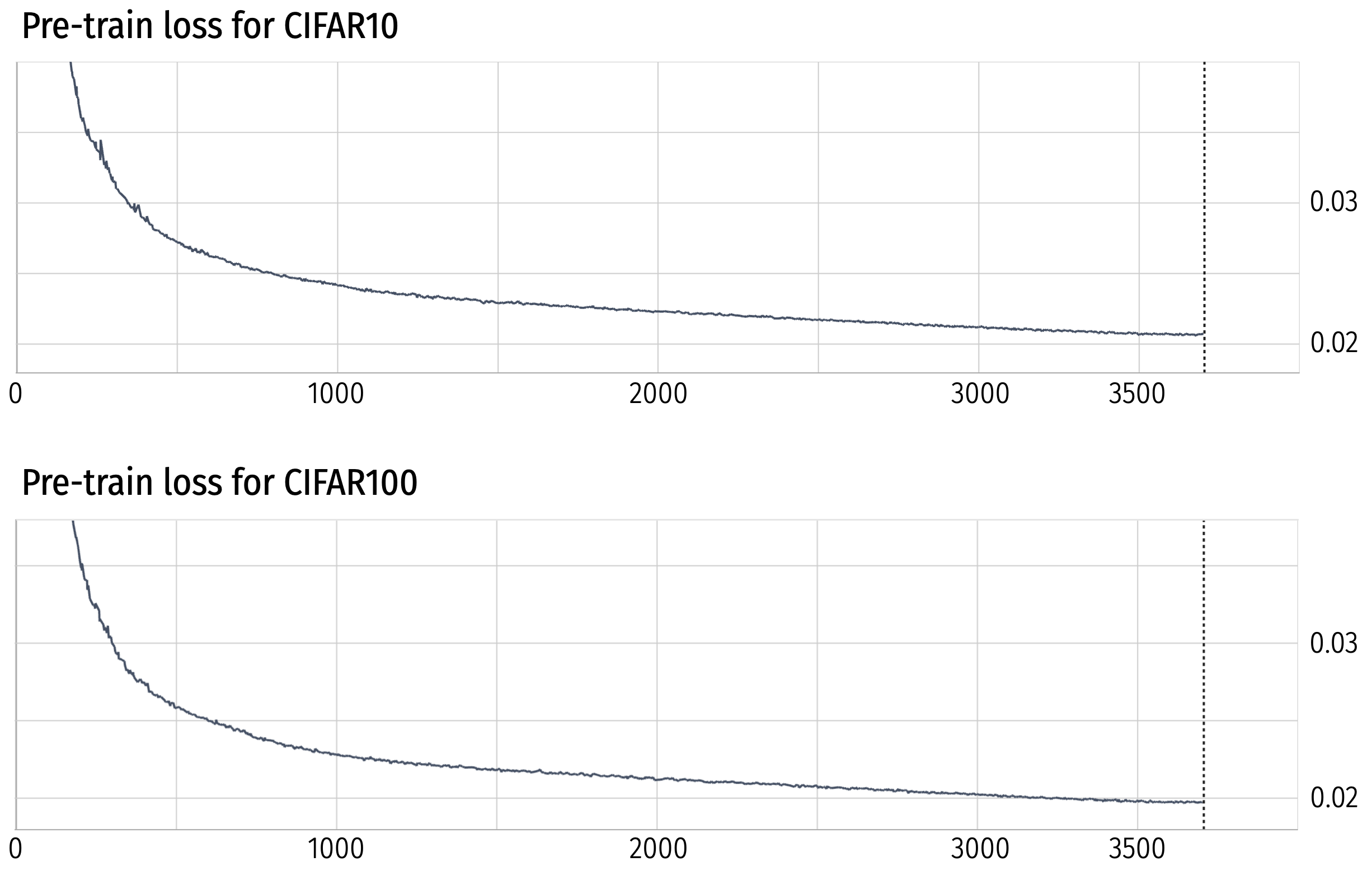}
    \caption{Pre-training loss for both CIFAR-10 and CIFAR-100 datasets. For CIFAR10, the training loss begins at 0.63 and concludes at 0.0207, while for CIFAR100, it starts at 0.66 and finishes at 0.0197. Notably, the y-axis of the plots is limited to the range of around 0.02 to 0.04 to enhance the visibility of the loss trends during most of the epochs. As a result of this scale adjustment, the higher training losses observed in the early epochs are not displayed in this figure. }
    \label{fig:mae_pretrain_loss}
\end{figure}

\subsection{On Pre-Training}

The models pre-trained on CIFAR-10 and CIFAR-100 are denoted as Mae-ViT-C10 and Mae-ViT-C100, respectively. Pre-training was scheduled for 4000 epochs; however, Mae-ViT-C10 concluded at 3700 epochs, and Mae-ViT-C100 ceased at 3703 epochs due to the end of training sessions. This early stopping was accepted because based on previous observations, there were no significant improvement in loss value noted in the last 300-400 epochs due to the very low learning rates at those epochs.

Pre-training on the more class-diverse CIFAR100 resulted in Mae-ViT-C100 achieving a marginally lower training loss of 0.019711, compared to 0.020733 for Mae-ViT-C10 (refer to Figure \ref{fig:mae_pretrain_loss}). While the difference of approximately 0.001 in training loss might seem minor, it is significant in the context of pre-training, where achieving such a reduction often requires hundreds of epochs at later stages.  As shown in Figures \ref{fig:mae_out} and \ref{fig:mae_stg}, the quality of the reconstructed images on visible patches remains high for both models, diverging from the original MAE results \cite{he2022masked} due to Eq. \ref{eq:loss}.

\begin{figure*} 
    \centering
    \includegraphics[trim=1.5cm 8cm 1.3cm 0cm, clip, scale=0.65]{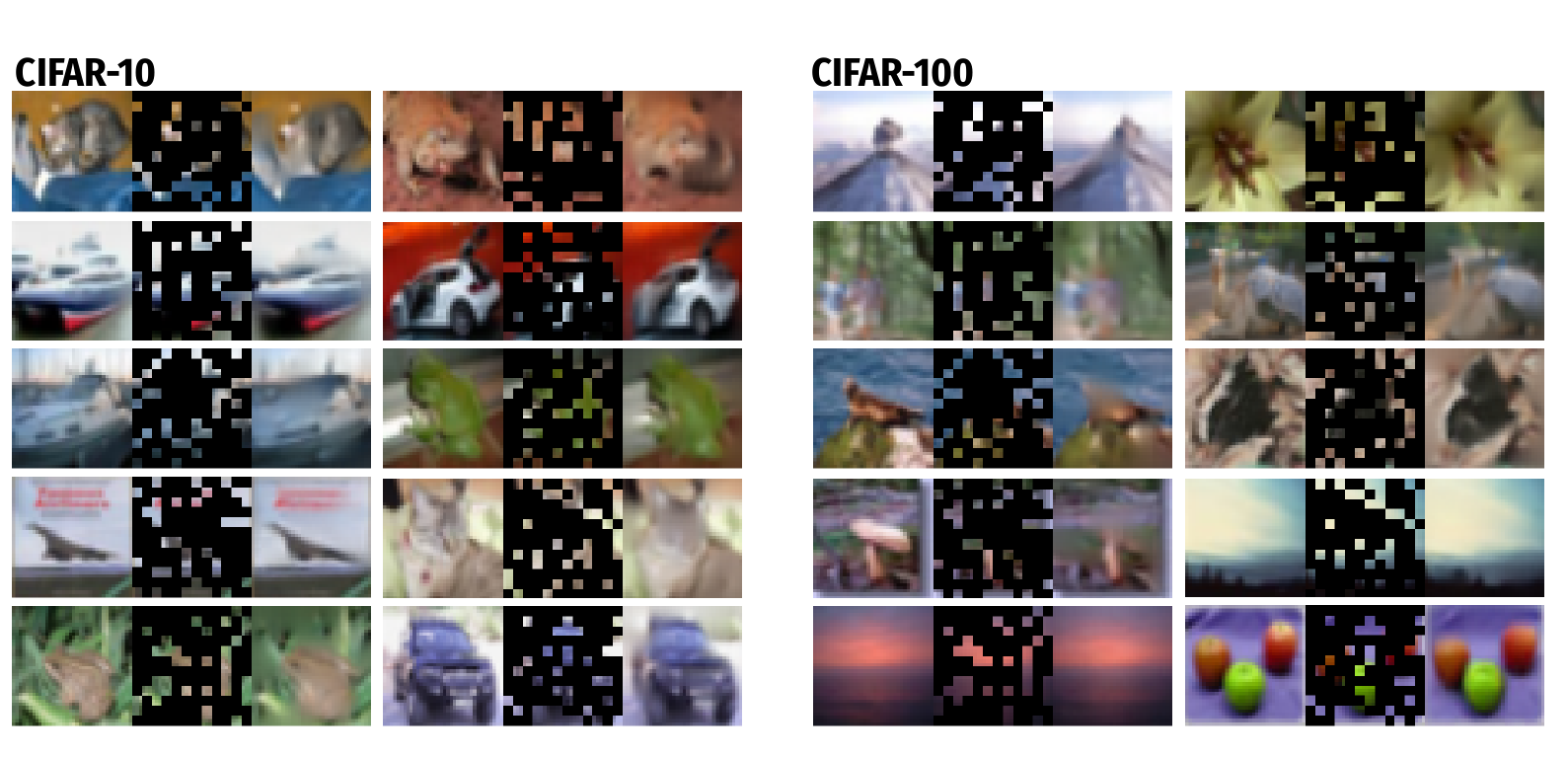}
    \caption{Example results on CIFAR10 and CIFAR100 \textit{validation} images. For each triplet, on the left is the original image. The middle is the masked image, and the right is the reconstructed image by MAE.}
    \label{fig:mae_out}
\end{figure*}

\subsection{On Fine-Tuning}
Each of the pre-trained models, Mae-ViT-C10 and Mae-ViT-C100, underwent fine-tuning for classification tasks on both CIFAR-10 and CIFAR-100 datasets. The performance results of these models are detailed in Table \ref{tab:result}. 

All of our models, which were first pre-trained and then fine-tuned, outperformed (in terms of accuracy) comparable transformer architectures that have similar counts of parameters and MACs. The highest accuracy achieved was \textbf{96.41\%} and \textbf{78.27\%} on CIFAR-10 and CIFAR-100 respectively. However, it was observed that models pre-trained and fine-tuned on the same dataset performed better than those pre-trained on one dataset and fine-tuned on the other. On the other hand, in the fine-tuning phase for CIFAR100, signs of overfitting appeared after 50 epochs, whereas this trend was less evident with CIFAR-10 (see Figure \ref{fig:mae_finetune}).

\begin{table}
\caption{Comparison of Top-1 validation accuracy}
\centering
\begin{tabular}{l|rr|rr}
\toprule
\textbf{Model} & \textbf{C10} & \textbf{C100} & \textbf{\# Params} & \textbf{MACs} \\
\midrule
\multicolumn{5}{l}{\textit{Convolutional Networks (Designed for CIFAR)}} \\
\midrule
\textbf{ResNet56} \cite{he2016deep} & 94.63\% & 74.81\% & 0.85 M & 0.13 G \\
\textbf{ResNet110}\cite{he2016deep} & 95.08\% & 76.63\% &  1.73 M & 0.26 G \\
\textbf{ResNet1k-v2*} \cite{he2016identity} & 95.38\% & --- & 10.33 M & 1.55 G \\
\midrule
\multicolumn{5}{l}{\textit{Vision Transformers}\cite{hassani2021escaping}} \\
\midrule
\textbf{ViT-12/16} & 83.04\% & 57.97\% &  85.63 M & 0.43 G \\
\midrule
\textbf{ViT-Lite-7/16} & 78.45\% & 52.87\% & 3.89 M & 0.02 G \\
\textbf{ViT-Lite-7/8} & 89.10\% & 67.27\% & 3.74 M & 0.06 G \\
\textbf{ViT-Lite-7/4} & 93.57\% & 73.94\% & 3.72 M & 0.26 G \\
\midrule
\multicolumn{5}{l}{\textit{Compact Vision Transformers}\cite{hassani2021escaping}} \\
\midrule
\textbf{CVT-7/8} & 89.79\% & 70.11\% & 3.74 M & 0.06 G \\
\textbf{CVT-7/4} & 94.01\% & 76.49\% & 3.72 M & 0.25 G \\
\midrule
\multicolumn{5}{l}{\textit{Compact Convolutional Transformers}\cite{hassani2021escaping}} \\
\midrule
\textbf{CCT-2/3 $\times$ 2} & 89.75\% & 66.93\% & 0.28 M & 0.04 G \\
\textbf{CCT-7/3 $\times$ 2} & 95.04\% & 77.72\% & 3.85 M & 0.29 G \\
\midrule
\multicolumn{5}{l}{\textit{MAE Vision Transformers} (\textbf{ours})} \\
\midrule
\textbf{Mae-ViT-C10} & \textbf{96.41\%} & 78.20\% & 3.64 M & 0.26 G \\
\textbf{Mae-ViT-C100} & 96.01\% & \textbf{78.27\%} & 3.64 M & 0.26 G \\
\bottomrule
\end{tabular}
\label{tab:result}
\end{table}

\begin{figure*} 
    \centering
    \includegraphics[trim=2cm 1cm 3cm 0cm, clip, scale=0.6]{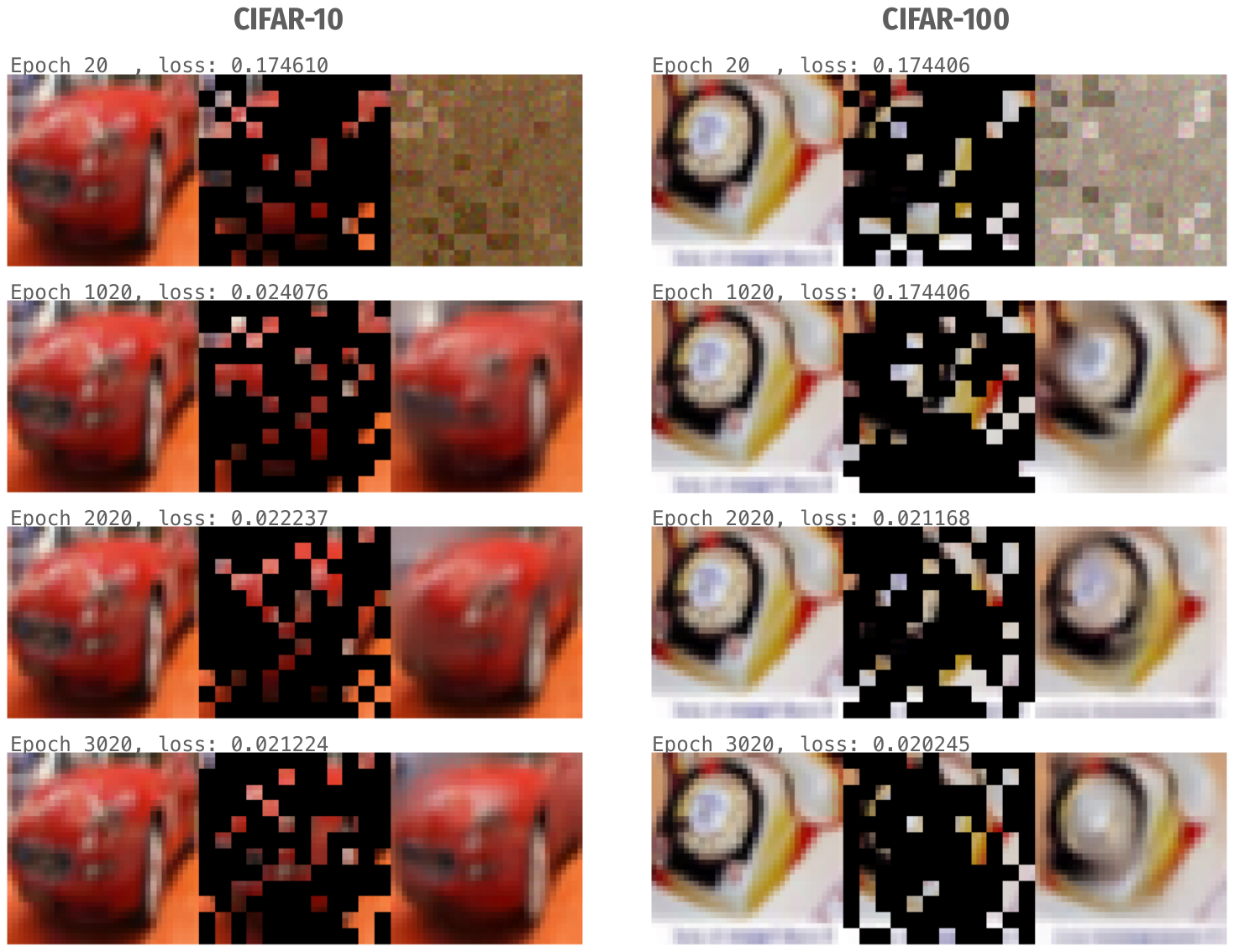}
    \caption{The evolution of the reconstructed outputs by the Masked Auto-Encoder at different training epochs: 20, 1020, 2020, and 3020. The left column displays a sample image from the CIFAR-10 training set, while the right column shows a corresponding sample from the CIFAR-100 training set. Each row corresponds to the reconstruction quality at the specified epoch, demonstrating the progressive refinement of the model's output over time.}
    \label{fig:mae_stg}
\end{figure*}

\begin{figure*} 
    \centering
    \includegraphics[trim=0cm 0cm 0cm 0cm, clip, scale=0.36]{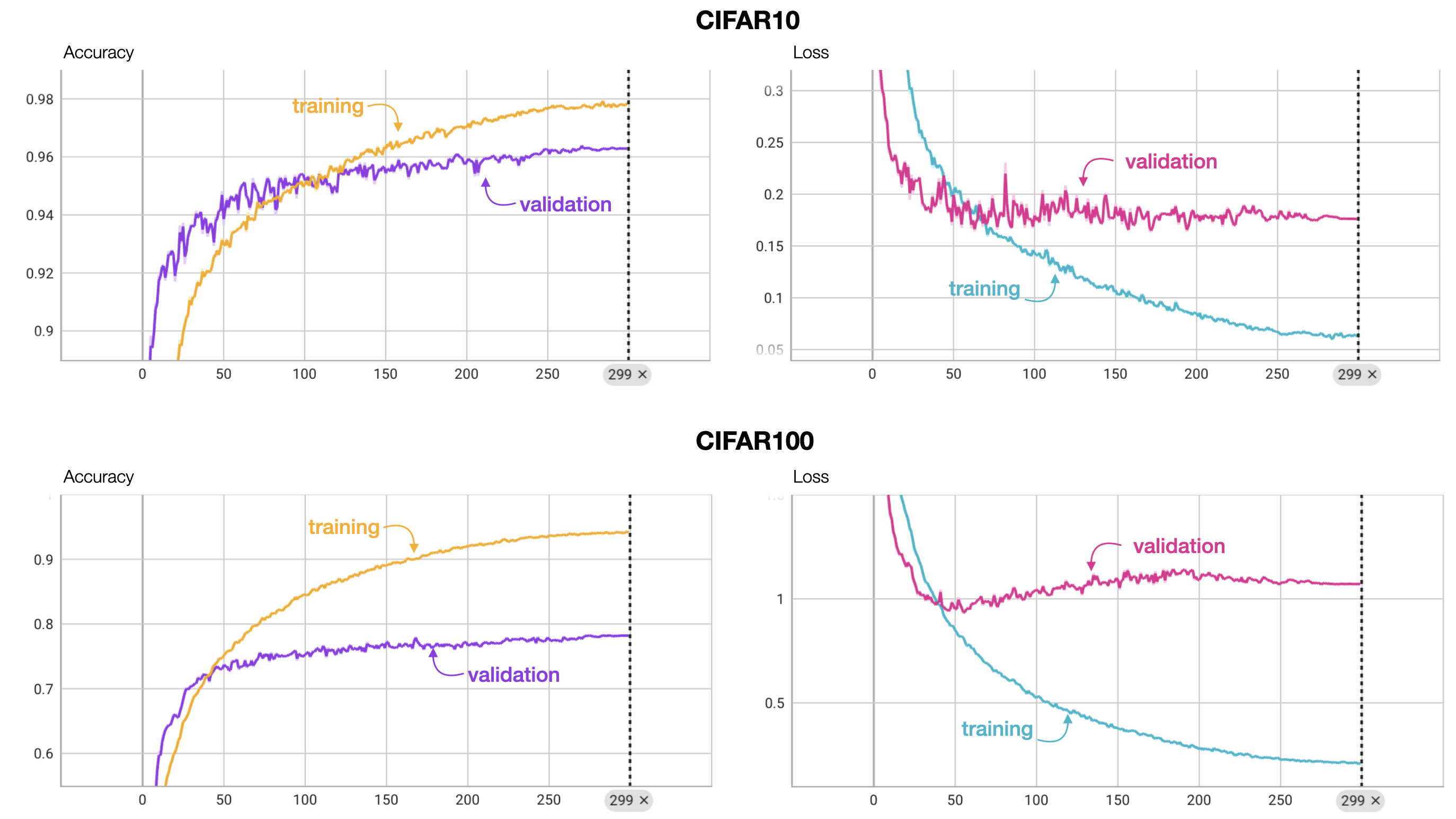}
    \caption{Comparison of training and validation performance for models subjected to both pre-training and fine-tuning on CIFAR-10 and CIFAR-100. The top row features plots of accuracy and loss for the model pre-trained on CIFAR-10 and subsequently fine-tuned on the same dataset. The bottom row follows a similar format, displaying the accuracy and loss for the model pre-trained on CIFAR-100 and fine-tuned on CIFAR-100. }
    \label{fig:mae_finetune}
\end{figure*}

\section{Discussion}
Table \ref{tab:result} reveals that Vision Transformers (ViT) without pre-training struggle to yield competitive results. It was reported that ViT struggled despite extended training epochs \cite{zhou2023self}. However, ViTs are sometimes favored due to their computational efficiency on hardware accelerators optimized for large matrix multiplications, which are central to transformer operations, as opposed to the more complex data access patterns required by convolutions \cite{graham2021levit}.

To enhance ViT competitiveness, it's common to integrate convolutional layers \cite{graham2021levit, hassani2021escaping} or employ a convolutional 'teacher' \cite{touvron2021training} for imparting inductive bias. Our report, however, demonstrates that with a masked auto-encoder setup, these measures are not necessary.

Our models notably outperform standard transformer architectures trained from scratch, as highlighted by the ViT-Lite-7/4 model (see Table \ref{tab:result}), which matches our models in parameter count and MACs. They also surpass variations of transformers that incorporate convolutional layers \cite{hassani2021escaping}, such as CVT-7/4 and CCT-7/3x2. Contrary to most state-of-the-art models that upscale input images (for CIFAR-10 and CIFAR-100), we minimally increase the image size to 36 x 36 from the original 32 x 32, proving the efficacy of vision transformers even with small datasets and resolutions.

The decoders in our lightweight pre-trained models, Mae-ViT-C10 and Mae-ViT-C100, produce satisfactory reconstructions. From our observations, we think a minimum learning rate in the cosine decay schedule could optimize the utility of the final few hundreds epochs, potentially benefiting a subsequent pre-training phase. A tentative minimum learning rate could be around 20\% of the value derived from Eq. \ref{eq:lr}.

MAE proves to be sample-efficient, learning robust representations with less pre-training data and minimizing overfitting compared to other methods \cite{el2024scalable}. While contrastive methods \cite{oquab2023dinov2,zhou2021image,caron2021emerging} often lead to stronger representations for the same model size, they face scalability challenges and loss tractability \cite{el2024scalable}. Pre-trained MAE models require fine-tuning for competitive performance, and while the resulting embeddings have lower linear separability, this does not necessarily correlate with transfer learning performance \cite{chen2021exploring}, and linear probing benchmarks are seldom used in natural language processing \cite{he2022masked}.

In this investigation, we find that models pre-trained on one dataset and fine-tuned on another did not outperform those pre-trained and fine-tuned on the same dataset. This could be due to the insufficient size in the secondary dataset. However, cross-dataset pre-training and fine-tuning did aid in faster convergence, despite the final performance being slightly lower than that of models trained and fine-tuned on the same dataset. In future investigations, it may be beneficial to explore pre-training on a merged dataset that includes both CIFAR-10 and CIFAR-100, followed by fine-tuning on each dataset separately.

\section{Conclusion}
This report has demonstrated that lightweight ViTs can indeed be pre-trained effectively on small datasets to achieve and even surpass the performance of traditional CNNs. By employing MAE, we have shown that the need for larger datasets or additional convolutional layers, which are commonly believed to be necessary for ViTs to be competitive, can be circumvented. 

\bibliographystyle{unsrt}  
\bibliography{references}

\end{sloppy}
\end{document}